\documentclass[acmtog,nonacm]{acmart}

\usepackage{mathtools}
\usepackage{graphicx}
\usepackage{booktabs}
\usepackage{multirow}
\usepackage{array}
\acmJournal{TOG}
\setcopyright{none}
\acmDOI{}
\acmISBN{}
\begin{document}

\title{LatentReRig: An SDF-Based VAE with Dual Decoders for Latent-Space Deformation Conditioning}

\author{Daniele Dolci}
\affiliation{%
  \institution{AnimSchool}
  \state{Utah}
  \country{USA}}
  \email{danieled@animschool.edu}
\affiliation{%
  \institution{Mostly Very True SRL}
  \city{Viterbo}
  \country{Italy}}
  \email{daniele@mvtpictures.com}
\author{Fabrizio Poggioni}
\affiliation{%
  \institution{Universit\`a degli Studi Guglielmo Marconi}
  \city{Rome}
  \country{Italy}}
\author{Carlo Melchiorri}
\affiliation{%
  \institution{Universit\`a degli Studi Guglielmo Marconi}
  \city{Rome}
  \country{Italy}}
  \email{c.melchiorri@unimarconi.it }
  
\renewcommand{\shortauthors}{Dolci et al.}

\begin{abstract}
Transferring deformation between characters with different geometry and
topology is challenging because conventional rigs encode behaviour through
character-specific structures and correspondences. We present LatentReRig,
an experimental framework that investigates whether pose-associated changes
can instead be represented as reusable directions in a learned geometric
latent space. An SDF-based variational autoencoder is coupled with two
decoders: one reconstructs the implicit field, while the other predicts
target vertex positions from source geometry and latent deformation
conditioning. The source geometry may be neutral or already deformed.
Experiments on a controlled humanoid dataset show that several poses induce
coherent latent directions across identities, particularly for broad
articulated motions. These signals can guide deformation of unseen
characters, but explicit predictions remain less accurate for localized
changes and corrective contributions. Diagnostic comparisons with repeated
SDF sampling show that inter-identity distances exceed same-geometry
resampling variability on average, while pose signals exhibit different
margins above this baseline. The results support the presence of reusable
pose-related structure and identify stable local conditioning and accurate
mesh decoding as complementary requirements for improving transfer.
\end{abstract}


\maketitle

\section{Introduction}
\label{sec:introduction}

Character rigging defines the relationship between a controllable representation of a character and the geometric deformations produced by that representation. In conventional production pipelines, such deformation behavior is typically encoded through a combination of skeletal transformations, skinning weights, corrective shapes, constraints, procedural systems, and artist-authored controls \citep{kavan2014skinning,lewis2000pose}. Although these mechanisms provide a high degree of control, the resulting deformation model is usually strongly dependent on the specific character for which it was created.

Transferring deformation behavior from one character to another therefore remains a difficult problem, particularly when the source and target identities differ substantially in proportions, morphology, or topology. Traditional approaches frequently rely on explicit geometric correspondences, common templates, non-rigid registration, wrapping procedures, or interpolation mechanisms \citep{sumner2004deformation,amberg2007nonrigid,ju2005mvc,joshi2007harmonic}. These requirements constrain the generality of the transfer process and make deformation reuse across heterogeneous characters particularly challenging.

This work investigates a different formulation of the problem. Rather than attempting to directly transfer vertex displacements, rig controls, or complete rigging structures, it asks whether deformation information can be represented within an automatically learned latent space and subsequently reused as a conditioning signal for the prediction of deformed geometry.

The proposed framework, \textit{LatentReRig}, explores this hypothesis using a variational representation \citep{kingma2014vae} learned from Signed Distance Function (SDF) samples of multiple character identities observed in both neutral and posed configurations. Signed-distance representations have long been used in implicit and level-set formulations \citep{osher1988fronts}, while neural SDF methods such as DeepSDF established their use as continuous learned shape representations \citep{park2019deepsdf}. The learned representation is coupled with two decoding objectives: an implicit SDF reconstruction task and an explicit mesh-space prediction task. This formulation makes it possible to investigate not only whether the latent representation can encode character geometry, but also whether identity-related and pose-related variations exhibit measurable and reusable structure.


\subsection{Research Problem}
\label{sec:research_problem}

The central research problem addressed in this work is the transfer of pose-associated geometric deformation between different character identities.

More specifically, given a deformation associated with a semantic pose on one or more source characters, the objective is to determine whether the information describing that deformation can be reused to condition the prediction of the corresponding pose on a previously unseen target identity.

Let an identity $i$ be represented in a neutral configuration by a geometry $X_i^0$, and let $X_i^p$ denote the geometry of the same identity under a semantic pose $p$. A conventional deformation representation could be expressed directly in vertex space as

\[
\Delta X_i^p = X_i^p - X_i^0.
\]

However, such a displacement is inherently dependent on the geometry, topology, vertex correspondence, and proportions of identity $i$. Consequently, the direct reuse of $\Delta X_i^p$ on another character $j$ is generally not well defined.

The problem considered here is therefore whether an alternative representation can be learned in which the effect of pose $p$ can be expressed independently enough from the identity on which it was observed to become reusable across characters.

If an encoder maps a character geometry to a latent representation $z$, the pose-dependent displacement for identity $i$ can instead be expressed as

\[
\Delta z_i^p = z_i^p - z_i^0,
\]

where $z_i^0$ and $z_i^p$ represent the neutral and posed latent encodings of the same identity.

The primary question is therefore not whether two characters undergo identical geometric displacements, which would generally be an unrealistic assumption, but whether the latent displacements induced by the same semantic pose exhibit sufficient cross-identity consistency to encode a reusable deformation signal.


\subsection{Central Hypothesis and Scope}
\label{sec:hypothesis}

The central hypothesis of this work is that character geometry admits an automatically learned, semantically organized latent representation in which identity-related and pose-related variations remain sufficiently distinguishable, and in which pose-induced latent displacements exhibit enough cross-identity consistency to be reused as conditioning signals for previously unseen characters.

Under this hypothesis, the latent displacement associated with the same semantic pose across two identities $i$ and $j$ should exhibit a degree of consistency such that

\[
\Delta z_i^p \approx \Delta z_j^p,
\]

not necessarily in absolute magnitude, but at least in terms of the direction and organization of the deformation represented within the latent space. A weaker and experimentally more appropriate expectation is therefore that

\[
\cos
\left(
\Delta z_i^p,
\Delta z_j^p
\right)
\rightarrow 1
\]

for character pairs affected by the same semantic pose.

This hypothesis does not require complete disentanglement between identity and pose. Rather, it assumes that the learned representation contains sufficient structure for identity-related variation and pose-related variation to be distinguished statistically and exploited computationally. Residual pose--identity entanglement is therefore treated as an experimental property of the learned space rather than as a violation of the problem formulation itself.

The scope of the work is deliberately restricted to \textit{pose-associated geometric deformation transfer}. LatentReRig does not attempt to reconstruct or transfer a complete production rig, infer an underlying skeletal hierarchy, recover artist-authored animation controls, reproduce procedural rigging logic, or perform general-purpose automatic rig generation.

Instead, the system investigates whether the geometric effect associated with a pose can be encoded in a sufficiently identity-independent form to condition the prediction of deformed geometry on a different character.

The experiments further assume a shared semantic pose vocabulary across identities. A pose label such as \textit{jawOpen}, \textit{elbowBent}, or \textit{cornerStretch}, for example, is assumed to identify an analogous deformation intent across different characters even when the resulting geometry, proportions, and local displacement magnitudes differ substantially.

Each identity is also assumed to provide a neutral reference configuration. This neutral state defines the identity-specific baseline from which pose-dependent latent variation can be measured. The objective is therefore not to remove identity information from the latent representation, but to determine whether deformation-related variation can be characterized relative to that identity baseline.


\subsection{Contributions}
\label{sec:contributions}

The main contributions of this work are:

\begin{itemize}

    \item an SDF-based variational framework for learning a common latent representation of multiple character identities and semantic poses;

    \item a dual-decoder architecture combining implicit SDF reconstruction with explicit mesh-space deformation prediction conditioned on source geometry and latent deformation information;

    \item an experimental formulation of pose-dependent latent displacements as differences between related geometric states, with neutral-to-pose transitions used to analyze cross-identity deformation consistency;

    \item a quantitative investigation of identity separation, pose consistency, and sensitivity to SDF resampling, including an empirical latent noise-floor analysis;

    \item an evaluation of the extent to which the learned deformation representation can condition explicit predictions on previously unseen character identities, together with an analysis of the principal failure modes and remaining pose--identity entanglement.

\end{itemize}


\section{Related Work}
\label{sec:related_work}

LatentReRig lies at the intersection of learned 3D representations,
latent-space modelling, deformation transfer, and neural character rigging.
The following review is intentionally selective and focuses on the lines of
work that directly motivate the proposed formulation.


\subsection{3D Geometric Representations and Implicit Fields}
\label{sec:shape_representations}

Three-dimensional geometry admits multiple computational representations,
each exposing different properties to a learning system. Classical solid
modelling already distinguished explicit boundary descriptions from implicit
and volumetric representations \citep{requicha1980solids}. In modern learning
pipelines, common choices include voxel grids, point clouds, polygonal meshes,
sparse spatial structures, and continuous implicit fields.

Voxel-based methods provide regular structures that are naturally compatible
with convolutional architectures. 3D ShapeNets and VoxNet demonstrated early
learned volumetric shape representations
\citep{wu2015shapenets,maturana2015voxnet}, but dense voxel grids scale
cubically with spatial resolution. Sparse hierarchical formulations such as
OctNet and O-CNN reduce this cost by allocating computation preferentially to
occupied or geometrically relevant regions
\citep{riegler2017octnet,wang2017ocnn}.

Point-cloud methods instead operate directly on sampled surface points.
PointNet showed that unordered point sets can be processed without converting
them to a dense volumetric grid \citep{qi2017pointnet}. Polygonal meshes retain
explicit surface connectivity and are compact and directly usable in
production, but varying topology and vertex count make them less immediately
compatible with fixed-size neural architectures.

Implicit representations describe geometry through a function evaluated over
continuous space. Signed-distance functions are closely related to level-set
representations, where a surface is represented as the zero level set of an
implicit scalar field \citep{osher1988fronts}. For a Signed Distance Function,
the scalar additionally measures distance to the closest surface with a sign
indicating the side of the surface.

Neural implicit representations brought this formulation into learned 3D
shape modelling. DeepSDF represents shape as a continuous signed-distance
function conditioned by a compact latent code \citep{park2019deepsdf}, while
Occupancy Networks learn a continuous inside--outside field rather than an
explicit distance \citep{mescheder2019occupancy}. These approaches avoid the
fixed spatial resolution of dense voxel grids and, importantly for the present
work, define their geometric input independently of the vertex indexing and
connectivity of the original polygonal mesh.

LatentReRig therefore uses sampled SDF observations to construct its implicit
representation, while retaining polygonal vertices only for the final explicit
deformation-prediction stage.


\subsection{Autoencoders and Latent Shape Spaces}
\label{sec:latent_spaces}

Representation learning seeks compact variables that preserve explanatory
structure in high-dimensional observations \citep{bengio2013representation}.
Autoencoders provide a direct formulation of this objective by learning an
encoder that maps an input into a latent code and a decoder that reconstructs
the original signal.

Variational Autoencoders (VAEs) replace a deterministic code with a learned
latent distribution and regularize that distribution toward a prior
\citep{kingma2014vae}. This produces a continuous probabilistic representation,
but also introduces an information-capacity trade-off. Related work has shown
that variational objectives can under-use latent dimensions or suppress
information that contributes little to reconstruction
\citep{burda2016iwae,alemi2018broken,higgins2017betavae}, and posterior-collapse
behaviour has been studied in several settings
\citep{bowman2016sentences,razavi2019deltavae}. These observations are relevant
to LatentReRig because small localized correctives may provide a much weaker
reconstruction signal than broad articulated changes.

The present work is specifically interested in whether semantic changes can
be represented as relative movements in latent space. The general idea that
semantic relationships may correspond to vector directions has a well-known
analogy in distributed word representations, where vector arithmetic can
capture regular semantic relationships \citep{mikolov2013word2vec}. LatentReRig
does not assume that geometric deformation obeys the same algebra exactly,
but adopts the related hypothesis that a pose can induce a repeatable latent
displacement across different identities.

Given the latent encodings of a neutral and posed state of identity $i$, the
pose-dependent displacement is

\begin{equation}
\Delta z_i^p = z_i^p - z_i^0.
\label{eq:related_latent_delta}
\end{equation}

If analogous poses generate sufficiently consistent latent variations across
identities, these differences may provide a deformation-conditioning signal
that is less directly tied to mesh topology than explicit vertex offsets.


\subsection{Deformation Transfer and Learned Deformation}
\label{sec:deformation_transfer}

Character deformation has traditionally been represented through skeletal
skinning and corrective mechanisms. Pose Space Deformation provides a
foundational example of pose-dependent corrective interpolation layered on
top of skeletal motion \citep{lewis2000pose}. Such methods are highly effective
when the rig and its deformation parameterization are already known, but do
not by themselves solve transfer between unrelated character identities.

A foundational formulation of mesh deformation transfer was introduced by
Sumner and Popovi\'c, who transfer local deformation gradients between
triangle meshes \citep{sumner2004deformation}. The method established an
important principle for the present work: transferring deformation is not
equivalent to copying vertex offsets, because the transferred transformation
must adapt to the target geometry.

When source and target shapes differ substantially, classical pipelines may
introduce non-rigid registration before transfer. Optimal Step Nonrigid ICP,
for example, estimates locally affine transformations under a stiffness
schedule \citep{amberg2007nonrigid}. Surface or cage relationships can then be
preserved through interpolation schemes such as Mean Value Coordinates
\citep{ju2005mvc} or Harmonic Coordinates, the latter specifically developed
for character articulation \citep{joshi2007harmonic}. These approaches are
powerful but depend on explicit correspondences, registration quality, cages,
or other transfer structures.

Learning-based methods instead infer deformation behaviour from examples.
Fast and Deep Deformation Approximations showed that neural networks can
efficiently approximate complex production deformation systems
\citep{bailey2018fast}. Neural Human Deformation Transfer addresses
cross-identity human deformation directly through a learned encoder--decoder
formulation \citep{basset2021neuralhuman}. Learning Skeletal Articulations with
Neural Blend Shapes learns pose-dependent corrective behaviour while retaining
an explicit skeletal animation model \citep{li2021neuralblendshapes}. VINECS
further demonstrates learned pose-dependent character skinning from multi-view
video \citep{liao2024vinecs}.

LatentReRig follows the general objective of learned deformation transfer but
investigates a different intermediate representation. Rather than transferring
a deformation through an explicit correspondence map or predicting it directly
from predefined skeletal parameters, the method first measures the deformation
as a difference between learned geometric latent codes and uses that difference
to condition an explicit mesh predictor.


\subsection{Neural Rigging and Related Systems}
\label{sec:neural_rigging}

A closely related research direction is automatic and neural rigging, where
the objective is to infer structures that can drive animation rather than only
predict a deformed mesh. Baran and Popovi\'c introduced an influential
geometry-driven automatic rigging method that adapts a predefined skeleton to
a static character mesh \citep{baran2007automatic}.

RigNet later formulated rigging as an end-to-end learning problem, predicting
joints, hierarchy, and skinning weights from character geometry
\citep{xu2020rignet}. TARig focuses on humanoid characters and introduces a
template-aware strategy intended to preserve a standardized skeletal structure
\citep{ma2023tarig}. HumanRig combines a large humanoid rigging dataset with
learned skeleton and skinning prediction \citep{chu2025humanrig}.

More recent systems broaden the range of assets that can be automatically
articulated. MagicArticulate predicts skeletons and skinning weights for static
3D assets \citep{song2025magicarticulate}; RigAnything formulates template-free
rig generation autoregressively for diverse object categories
\citep{liu2025riganything}; and UniRig targets a unified rigging model using
large-scale autoregressive modelling and bone--point interaction
\citep{zhang2025unirig}. CANRig addresses a more specialized but production-
relevant domain by learning facial rigging with controllable local regions
\citep{mohammadi2026canrig}.

These systems differ from LatentReRig in their primary output. Neural rigging
typically predicts an explicit animation structure---for example joints,
hierarchy, skinning weights, or facial controls. LatentReRig instead treats the
deformed geometry generated by existing rigs as the observation to be learned.
Its distinguishing question is whether differences between latent encodings of
these rig-generated mesh states can become reusable conditioning signals for
explicit deformation prediction on previously unseen identities.


\section{LatentReRig Method}
\label{sec:method}


\subsection{Method Overview}
\label{sec:method_overview}

LatentReRig is designed as a hybrid implicit--explicit deformation framework.
Its objective is to learn a geometric latent representation in which variations
associated with character identity and pose can be measured, and subsequently
use these variations to condition explicit mesh deformation prediction.

\begin{figure}[t]
    \centering
    \includegraphics[width=\linewidth]{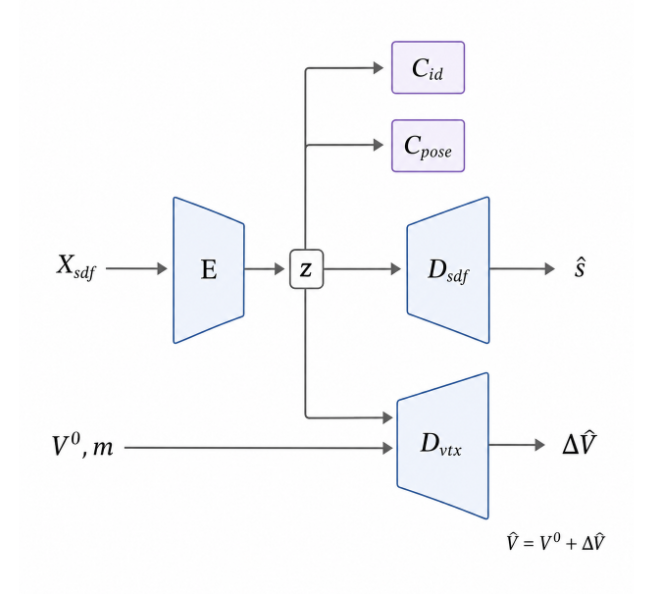}
    \caption{
        High-level overview of LatentReRig. SDF samples are encoded into a
        variational latent representation used for implicit SDF reconstruction
        and semantic regularization. Relative latent variation subsequently
        conditions an explicit vertex decoder for mesh-space deformation
        prediction. Lower-level decoder inputs are omitted for clarity.
    }
    \Description{Pipeline diagram connecting SDF encoding, implicit reconstruction, latent deformation conditioning, and explicit vertex prediction.}
    \label{fig:architecture_overview}
\end{figure}

The complete pipeline is illustrated in
Figure~\ref{fig:architecture_overview}. Rig-generated character states are
first converted into Signed Distance Function (SDF) samples and encoded by a
variational encoder. The resulting latent representation is used by two
different decoding branches. An SDF decoder reconstructs the implicit geometry
and provides geometric pressure on the latent space, while a vertex decoder
uses latent deformation information together with explicit source geometry to
predict target vertex positions.

The overall processing chain can therefore be summarized as

\begin{equation}
\begin{aligned}
\text{rig-generated geometry}
&\rightarrow
\text{SDF representation}
\\
&\rightarrow
\text{variational latent space}
\\
&\rightarrow
\Delta\mu
\rightarrow
\text{vertex conditioning}
\\
&\rightarrow
\hat{V}_{\mathrm{target}}.
\end{aligned}
\end{equation}

The two decoders have deliberately different roles. The SDF decoder determines
whether the latent representation preserves sufficient information to describe
the underlying geometry, whereas the vertex decoder determines whether
relative variations in that representation can be translated back into an
explicit deformation.

\subsection{Dataset and Geometric Representation}
\label{sec:dataset}

The dataset is generated from production-oriented humanoid character rigs in
Autodesk Maya. Rather than collecting arbitrary mesh states, each character is
evaluated according to a controlled deformation vocabulary containing neutral
configurations, isolated poses, combined poses, and different deformation-stack
variants.

The final geometry corpus contains 78 humanoid biped identities. Characters
are divided at identity level into 64 training characters and 14 test
characters, such that the test identities are never observed during training.
The complete dataset contains 31,794 mesh samples.

For posed configurations, three deformation variants are exported:
\textit{skinning-only}, \textit{blendshape-only}, and
\textit{skinning-plus-blendshape}. The dataset contains 10,572 pose
configurations, producing

\[
10{,}572 \times 3 = 31{,}716
\]

posed meshes. Including one neutral state for each of the 78 identities gives

\[
31{,}716 + 78 = 31{,}794
\]

total mesh states.

This organization exposes the network to three complementary sources of
variation: identity, semantic pose, and deformation regime. Single poses
represent isolated deformation concepts, while combined poses expose
interactions between multiple deformation sources. The deformation variants
separate coarse skeletal motion from corrective and final combined
deformation.

Each exported mesh state is represented in two complementary forms. The first
is an implicit SDF representation composed of spatial query points and their
signed distance values,

\[
X_{\mathrm{SDF}}
=
\left\{
(\mathbf{x}_i,s_i)
\right\}_{i=1}^{K},
\]

where $\mathbf{x}_i\in\mathbb{R}^{3}$ and $s_i\in\mathbb{R}$.

The second representation contains explicit vertex positions and surface
normals. These explicit data are organized into source--target pairs and are
used by the vertex decoder during deformation training.

The SDF and vertex datasets therefore describe different views of the same
rig-generated states. The SDF representation provides a geometry encoding
that is not directly defined by polygon connectivity, while the vertex data
provide the explicit supervision required to reconstruct usable deformed
meshes.


\subsection{SDF-Based Variational Representation}
\label{sec:sdf_vae}

The implicit branch consists of an SDF encoder and an SDF decoder. The use of
continuous signed-distance observations follows the neural implicit
representation established by DeepSDF \citep{park2019deepsdf}, while the
probabilistic latent formulation follows the VAE framework
\citep{kingma2014vae}. Given a set of sampled SDF observations, the encoder
produces the parameters of a variational latent distribution,

\[
(\mu,\log\sigma^{2})
=
E_{\phi}(X_{\mathrm{SDF}}),
\]

with latent dimensionality

\[
Z = 512.
\]

During variational training, a latent sample is obtained using the
reparameterization trick \citep{kingma2014vae},

\[
z
=
\mu
+
\sigma \odot \epsilon,
\qquad
\epsilon \sim \mathcal{N}(0,I).
\]

For latent analysis and deformation conditioning, LatentReRig uses the
posterior mean $\mu$ rather than a stochastic sample. This distinction is important because the
deformation representation is constructed from differences between latent
codes and should not depend unnecessarily on variational resampling.

The SDF decoder is a conditional implicit function. Given a latent code and a
query position, it predicts the corresponding signed distance,

\[
\hat{s}
=
D_{\psi}(z,\mathbf{x}).
\]

The latent code describes the character state, while the query coordinate
specifies the spatial location at which the field is evaluated. Reconstructing
the SDF forces the latent representation to preserve geometric information
about the encoded character and pose state.

Two auxiliary classification heads are used during this representation-learning
stage. A character classifier predicts the training identity from the latent
representation, while a pose classifier predicts the semantic pose from a
normalized pose-related latent difference. These heads are not used as final
classification systems. Their purpose is to introduce supervised semantic
pressure on the representation: identity information should remain accessible
from the latent code, while pose information should remain accessible from
relative latent variation.

Importantly, these objectives encourage but do not establish semantic
disentanglement. Whether the resulting space actually exhibits consistent
identity and pose structure is evaluated experimentally rather than assumed
from the optimization objectives.


\subsection{Latent Deformation Conditioning}
\label{sec:latent_conditioning}

The central representation used for deformation transfer is the difference
between the deterministic latent means of two related mesh states. Given a
source state and a target state,

\[
\mu_{\mathrm{source}}
=
E_{\phi}^{\mu}(X_{\mathrm{source}})
\]

and

\[
\mu_{\mathrm{target}}
=
E_{\phi}^{\mu}(X_{\mathrm{target}}),
\]

the raw latent deformation difference is

\[
\Delta \mu_{\mathrm{raw}}
=
\mu_{\mathrm{target}}
-
\mu_{\mathrm{source}}.
\]

This representation is decomposed into magnitude and normalized direction.
The magnitude is

\[
r_{\Delta}
=
\left\|
\Delta \mu_{\mathrm{raw}}
\right\|,
\]

while the normalized direction is

\[
\overline{\Delta \mu}
=
\frac{
\Delta \mu_{\mathrm{raw}}
}{
\left\|
\Delta \mu_{\mathrm{raw}}
\right\|
+
\varepsilon_{\mathrm{norm}}
}.
\]

The separation is useful because two instances of the same semantic
deformation may exhibit similar latent directions while differing in
magnitude. This use of relative latent directions is conceptually related to
vector regularities observed in distributed semantic representations
\citep{mikolov2013word2vec}, although no exact linear analogy is assumed here. Direction therefore describes the organization of the latent
transition, whereas magnitude provides information about the scale of that
transition.

A conceptual latent-transfer model could be expressed as

\[
\mu_{j}^{p}
\approx
\mu_{j}^{0}
+
\Delta \mu^{p},
\]

but LatentReRig does not directly decode a mesh from this addition. Instead,
the latent difference is used as a conditioning signal for an explicit vertex
decoder. This allows deformation information to be interpreted in the context
of the geometry of the character being deformed rather than assuming that a
single latent translation can directly generate the target surface.


\subsection{Explicit Vertex Decoder}
\label{sec:vertex_decoder}

The vertex decoder converts latent deformation information into explicit
mesh-space predictions. For a source--target training pair, its conceptual
input--output relation is

\[
\left(
V_{\mathrm{source}},
N_{\mathrm{source}},
\mu_{\mathrm{source}},
\overline{\Delta \mu},
r_{\Delta},
k
\right)
\rightarrow
\hat{V}_{\mathrm{target}},
\]

where

\begin{itemize}
    \item $V_{\mathrm{source}}$ contains the source vertex positions;
    \item $N_{\mathrm{source}}$ contains the corresponding surface normals;
    \item $\mu_{\mathrm{source}}$ represents the source identity and geometric state;
    \item $\overline{\Delta \mu}$ is the normalized latent deformation direction;
    \item $r_{\Delta}$ is its magnitude; and
    \item $k$ identifies the source--target pair kind.
\end{itemize}

The target vertex positions are used as geometric supervision rather than
as direct decoder inputs. During training, the target state also supplies
the latent mean used to construct the conditioning delta. At transfer time,
this delta is obtained from source identities as described in
Section~\ref{sec:mesh_prediction}.

The decoder combines global latent conditioning with local geometric
information. Its architecture follows a coarse-to-fine formulation composed
of five principal elements: a virtual-bone coarse stage, a per-vertex latent
delta mask, four local refinement stages, an optional residual detail branch,
and pair-kind conditioning.

The virtual-bone stage predicts a coarse articulated deformation through a
set of learned virtual influences. These virtual structures are not intended
to recover the production skeleton of the source character; they provide an
internal mechanism for approximating large-scale coordinated vertex motion.

The per-vertex latent mask allows a global deformation signal to be interpreted
differently across the surface. This is necessary because a semantic
deformation is generally localized: for example, a jaw deformation should
produce a stronger response around the lower face than on unrelated body
regions.

Subsequent refinement stages operate using local K-nearest-neighbour
relationships between vertices. They progressively correct the coarse
prediction while preserving local geometric context. A residual detail branch
provides additional capacity for small and localized deformation components,
which are particularly relevant for corrective-shape prediction.

Finally, pair-kind conditioning informs the decoder about the semantic
relationship between source and target states. Pair types include broad
neutral-to-pose transitions as well as deformation-regime transitions such as

\texttt{skinning\_\allowbreak to\_\allowbreak skinning\_\allowbreak plus\_\allowbreak blendshape}.

This distinction allows the same latent representation to be interpreted
differently depending on whether the requested transformation represents a
large pose deformation, a localized corrective contribution, or a composed
target state.


\subsection{Learning Objectives}
\label{sec:losses}

The optimization objectives reflect the two different roles of the
architecture: representation learning in the implicit branch and explicit
deformation prediction in the vertex branch.

During SDF representation training, the primary objective combines weighted
SDF reconstruction, character classification, pose classification, and
variational regularization,

\[
\mathcal{L}_{\mathrm{SDF}}
=
\lambda_{\mathrm{rec}}
\mathcal{L}_{\mathrm{rec}}
+
\lambda_{\mathrm{cls}}
\mathcal{L}_{\mathrm{cls}}
+
\lambda_{\mathrm{pose-cls}}
\mathcal{L}_{\mathrm{pose-cls}}
+
\lambda_{\mathrm{KL}}
\mathcal{L}_{\mathrm{KL}}.
\]

The reconstruction term forces the latent representation to preserve implicit
geometry. The identity and pose classification terms introduce semantic
supervision, while the KL divergence regularizes the variational distribution
toward a standard normal prior,

\[
p(z)=\mathcal{N}(0,I).
\]

The KL term is defined as

\[
\mathcal{L}_{\mathrm{KL}}
=
-\frac{1}{2}
\operatorname{mean}
\left(
1+\log\sigma^{2}
-\mu^{2}
-\exp(\log\sigma^{2})
\right).
\]

Additional latent-space regularization is introduced to encourage the
representation required by the central hypothesis. In particular, the model
is encouraged to produce consistent latent directions for related
deformations, to preserve approximately additive relationships between
compatible deformation variants and combined poses, and to separate latent
directions associated with different semantic poses.

These objectives are grouped as

\begin{equation}
\begin{aligned}
\mathcal{L}_{\mathrm{latent}}
&=
\lambda_{\mathrm{variant}}
\mathcal{L}_{\mathrm{variant-comp}}
+
\lambda_{\mathrm{combo}}
\mathcal{L}_{\mathrm{combo-comp}}
\\
&\quad+
\lambda_{\mathrm{contrast}}
\mathcal{L}_{\mathrm{contrast}}.
\end{aligned}
\end{equation}

For example, when compatible pose components $A$ and $B$ and their combined
configuration are available, the desired relation is

\[
\Delta \mu_{A+B}
\approx
\Delta \mu_A
+
\Delta \mu_B.
\]

Similarly, the contrastive objective encourages latent deltas associated with
the same pose to align while preventing different pose classes from
collapsing onto the same direction. These terms should be understood as
hypothesis-driven regularization; their presence does not imply that the
resulting latent space actually satisfies these properties.

The vertex-decoder objective operates directly on explicit geometry. Rather
than relying on a single full-mesh position error, it combines complementary
terms measuring target accuracy, local neighbourhood consistency, static
region preservation, displacement smoothness, and surface-normal agreement,

\begin{equation}
\begin{aligned}
\mathcal{L}_{\mathrm{vertex}}
&=
\lambda_{\mathrm{pos}}
\mathcal{L}_{\mathrm{pos}}
+
\lambda_{\mathrm{knn}}
\mathcal{L}_{\mathrm{knn}}
\\
&\quad+
\lambda_{\mathrm{knn-delta}}
\mathcal{L}_{\mathrm{knn-delta}}
+
\lambda_{\mathrm{static}}
\mathcal{L}_{\mathrm{static}}
\\
&\quad+
\lambda_{\mathrm{smooth}}
\mathcal{L}_{\mathrm{smooth}}
+
\lambda_{\mathrm{nrm}}
\mathcal{L}_{\mathrm{nrm}}.
\end{aligned}
\end{equation}

For corrective pairs, an additional specialized term emphasizes the
difference between a skinning-only state and its corresponding
skinning-plus-blendshape state,

\[
\Delta V_{\mathrm{corr}}^{\mathrm{gt}}
=
V_{\mathrm{skinning+blendshape}}
-
V_{\mathrm{skinning}}.
\]

The corrective objective is therefore

\[
\mathcal{L}_{\mathrm{vertex-corr}}
=
\mathcal{L}_{\mathrm{vertex}}
+
\lambda_{\mathrm{skinBS}}
\mathcal{L}_{\mathrm{skinBS}}.
\]

This additional weighting is necessary because corrective deformations may
affect only a small fraction of the surface and can otherwise be numerically
dominated by the large number of vertices that are expected to remain static.


\subsection{Staged Training Strategy}
\label{sec:training}

Training is divided into two sequential phases in order to decouple latent
representation learning from explicit deformation prediction.

\textbf{Phase A --- Representation learning.}
The SDF encoder, SDF decoder, character classifier, and pose classifier are
trained from SDF samples. The objective of this phase is to learn a latent
space that preserves geometric information while being exposed to semantic
and latent-structure regularization. The vertex decoder is inactive during
this phase.

The active objective can be summarized as

\[
\mathcal{L}_{A}
=
\mathcal{L}_{\mathrm{SDF}}
+
\mathcal{L}_{\mathrm{latent}}.
\]

\textbf{Phase B --- Explicit deformation learning.}
After the representation has been learned, the SDF encoder is frozen and
placed in evaluation mode. The SDF decoder and classification heads are no
longer optimized, while the vertex decoder is trained from explicit
source--target mesh pairs.

For each pair, the frozen encoder produces deterministic source and target
latent means,

\[
\mu_{\mathrm{source}},
\qquad
\mu_{\mathrm{target}},
\]

from which the deformation condition is constructed as

\[
\Delta \mu_{\mathrm{raw}}
=
\mu_{\mathrm{target}}
-
\mu_{\mathrm{source}}.
\]

The vertex decoder then predicts

\[
\begin{aligned}
\hat{V}_{\mathrm{target}}=D_{\mathrm{vert}}\bigl(&V_{\mathrm{source}},
N_{\mathrm{source}},\mu_{\mathrm{source}},\\
&\overline{\Delta\mu},r_{\Delta},k\bigr),
\end{aligned}
\]

and only the vertex-decoder parameters are updated using the explicit
geometric objectives.

Freezing the encoder during Phase B is methodologically important. If the
latent representation continued to change while the vertex decoder was being
optimized, the meaning of its conditioning variables would also move
throughout training. The staged procedure instead establishes the latent
representation first and then evaluates whether a separate explicit decoder
can learn to exploit the structure already present in that representation.

This separation also provides a useful experimental distinction. Phase A
tests whether a semantically useful geometric latent representation can be
learned, whereas Phase B tests whether that representation contains sufficient
information to support explicit deformation prediction.


\section{Experimental Evaluation}
\label{sec:experiments}

The experimental evaluation investigates the two successive requirements
underlying the LatentReRig hypothesis. First, the learned representation must
contain deformation-related structure that remains sufficiently consistent across
different character identities. Second, this structure must be informative enough to
condition explicit geometric prediction on characters not used to derive the
corresponding deformation signal.

The evaluation is therefore not reduced to a single reconstruction metric.
Character deformation is inherently multi-objective: a useful prediction must move
the intended region, preserve regions that should remain static, maintain plausible
local surface structure, and reproduce the semantic intent of the target deformation.
For this reason, quantitative latent-space measurements are considered together with
explicit mesh predictions and visual inspection in Maya.

The diagnostic analysis of SDF-resampling variability and the latent noise floor is
reported separately in Section~\ref{sec:latent_analysis}, while systematic failure
modes are discussed in Section~\ref{sec:limitations}.


\subsection{Evaluation Setup}
\label{sec:evaluation_setup}

The evaluation uses the identity-level split introduced in
Section~\ref{sec:dataset}. Characters assigned to the test set are excluded from
training, allowing the explicit deformation stage to be evaluated on previously
unseen identities.

The experimental corpus contains multiple deformation relationships. The principal
pair kinds are:

\begin{itemize}

    \item \texttt{neutral\_\allowbreak to\_\allowbreak pose}, in which a neutral character is transformed
    toward a posed state;

    \item \texttt{skinning\_\allowbreak to\_\allowbreak skinning\_\allowbreak plus\_\allowbreak blendshape}, in which the model
    predicts the corrective contribution required on top of an existing skinned pose;

    \item \texttt{blendshape\_\allowbreak to\_\allowbreak skinning\_\allowbreak plus\_\allowbreak blendshape}, in which an existing
    blendshape deformation is combined with the final deformation state; and

    \item \texttt{source\_\allowbreak to\_\allowbreak combo}, in which compatible deformation components
    condition the prediction of a combined target configuration.

\end{itemize}

These pair kinds represent deformation regimes of substantially different scale.
Neutral-to-pose examples often involve broad articulated motion affecting large
regions of the character, whereas corrective transitions may consist of small,
localized changes affecting only a limited subset of vertices.

This distinction is important because a weak prediction may result either from an
insufficiently represented latent condition or from limitations of the explicit vertex
decoder. The experiments therefore consider representation-level and
deformation-level evidence separately.

The principal evaluation criteria include SDF reconstruction behaviour, explicit
vertex prediction, movement of the intended deformation region, preservation of
static regions, local surface coherence, normal consistency, and cross-identity
latent-delta consistency.


\subsection{Latent Representation Evaluation}
\label{sec:representation_results}

The first requirement of the proposed method is that the SDF-based variational
representation retains meaningful information about the encoded character states.
During Phase A, the encoder is optimized jointly with the SDF decoder and the
auxiliary identity- and pose-related objectives described in
Section~\ref{sec:losses}.

The SDF reconstruction objective establishes that the latent representation contains
sufficient geometric information to support implicit reconstruction. However,
reconstruction quality alone is not evidence of a useful deformation space. An
autoencoder may accurately reconstruct different states while arranging them in a
latent representation whose relative differences have little semantic meaning.

For this reason, representation quality is evaluated primarily through the behaviour
of relative latent codes. The relevant quantity for a pose $p$ and identity $i$ is

\begin{equation}
\Delta \mu_i^p
=
\mu_i^p
-
\mu_i^0,
\label{eq:eval_pose_delta}
\end{equation}

where $\mu_i^0$ and $\mu_i^p$ are the deterministic latent means of the neutral and
posed states respectively.

The central experimental question is therefore whether the same semantic pose
produces similar latent transitions across different identities, rather than whether
the absolute latent codes themselves coincide.


\subsection{Cross-Identity Pose Consistency}
\label{sec:pose_consistency}

Cross-identity consistency is measured by comparing normalized pose deltas.
For identities $i$ and $j$,

\begin{equation}
c_{ij}^{p}
=
\frac{
\Delta \mu_i^p \cdot \Delta \mu_j^p
}{
\left\|\Delta \mu_i^p\right\|
\left\|\Delta \mu_j^p\right\|
}.
\label{eq:pose_cosine}
\end{equation}

A value approaching one indicates that the same pose induces a similar latent
direction for the two identities. The most consistent group,
\textit{elbowBentLf}, reaches a mean cosine similarity of approximately
$0.958\pm0.030$. Several other groups, including
\textit{fingerKnuckleD03Lf}, \textit{cornerPuckerLf},
\textit{clavicleDownwardLf}, and \textit{fingerKnuckleD02Lf}, exceed $0.93$.

These strong transitions also tend to have comparatively large latent magnitudes.
For example, \textit{elbowBentLf} has a mean norm of approximately $0.0298$,
while \textit{clavicleDownwardLf} has a mean norm of approximately $0.0236$
and mean cosine similarity of approximately $0.932$.

Localized deformations are less stable. \textit{browDwLf} has a mean norm of
approximately $0.0106$ and mean cosine similarity of $0.777$, while
\textit{puffLf} reaches approximately $0.743\pm0.306$.

The representation therefore does not encode all poses equally. Broad or
structurally constrained changes tend to generate more coherent cross-identity
directions, while small facial, finger, and corrective-style signals are more
fragile.

\subsection{Explicit Mesh Prediction on Unseen Identities}
\label{sec:mesh_prediction}

The second experimental question is whether latent deformation structure can be
translated into explicit geometry on identities excluded from training. Two inference
strategies are compared.

The first, \texttt{mean\_\allowbreak pose\_\allowbreak delta}, averages the latent transitions associated
with pose $p$,

\begin{equation}
\Delta \mu_{\mathrm{mean}}^p
=
\frac{1}{N_p}
\sum_{i=1}^{N_p}
\Delta \mu_i^p.
\label{eq:mean_pose_delta}
\end{equation}

This strategy represents the strongest interpretation of an identity-independent
pose direction. The second, \texttt{nearest\_\allowbreak identity\_\allowbreak delta}, uses the deformation
delta associated with an identity closest to the target in latent space, retaining
more character-specific information.

The two strategies produce visibly different results and neither dominates across
all poses. Broad limb and clavicle motions frequently reproduce the principal
semantic direction of the requested deformation, although the final surface remains
less accurate than the artist-authored target. Small finger, facial, and
corrective-style changes are less reliable and can become weak or incomplete.

The occasional advantage of nearest-identity conditioning indicates that a reusable
semantic component is present, but is not fully independent of character morphology.
This residual identity dependence is analyzed further in
Section~\ref{sec:identity_pose_entanglement} and discussed in
Section~\ref{sec:limitations}.

\begin{figure}[t]
    \centering
    \includegraphics[width=\linewidth]{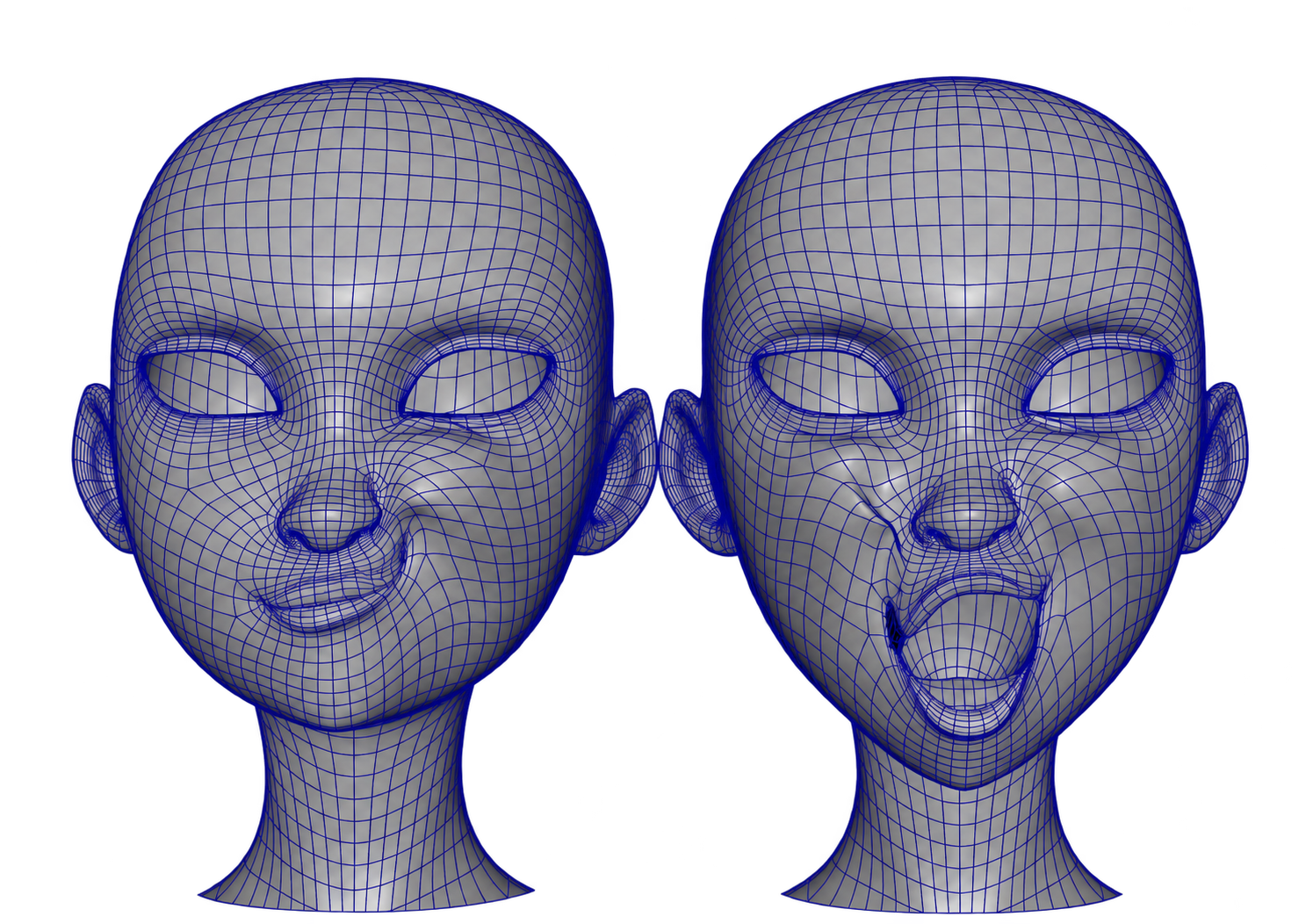}
    \caption{Effect of the conditioning strategy on \textit{cornerUpLf} for the held-out character.
    Left: nearest-identity delta. Right: mean pose delta across identities.
    Nearest-identity conditioning yields a more localized and visually coherent
    mouth-corner deformation in this example; the global mean produces a
    pronounced unintended mouth opening. Both meshes are predictions.}
    \Description{Two frontal wireframe face predictions. The left has a localized raised mouth corner; the right shows substantial mouth opening.}
    \label{fig:corner_conditioning}
\end{figure}

Figure~\ref{fig:corner_conditioning} illustrates this dependence for
\textit{cornerUpLf} on the held-out character: the nearest-identity condition produces a more
localized result, whereas the global mean introduces substantial mouth distortion.
This is a qualitative comparison of the two predictions, rather than a
measurement of their error against a displayed ground-truth mesh.

\subsection{Application-Level Case Study}
\label{sec:case_study}

LatentReRig was integrated into a lightweight Autodesk Maya prototype and tested
on a stylized production character excluded from training. The
prototype accepts a Maya mesh, exports and normalizes the geometry, performs SDF
sampling and latent encoding, selects a deformation condition, evaluates the vertex
decoder, and reconstructs the resulting positions as a new mesh.

\begin{figure*}[t]
    \centering
    \begin{minipage}[b]{0.48\textwidth}
        \centering
        \includegraphics[height=5.3cm,width=\linewidth,keepaspectratio]{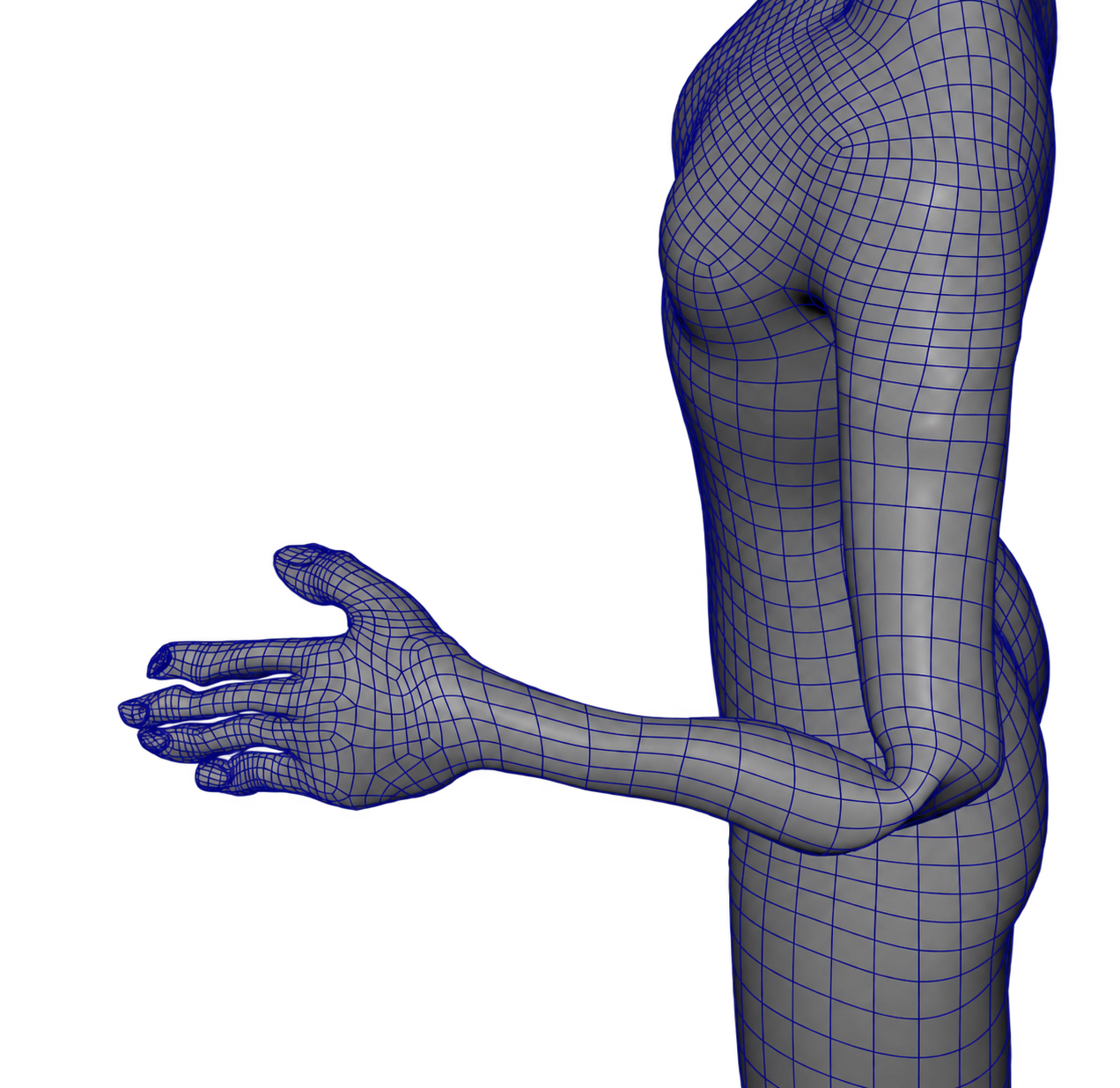}\\[3pt]
        \small (a) \textit{elbowBentLf}
    \end{minipage}\hfill
    \begin{minipage}[b]{0.48\textwidth}
        \centering
        \includegraphics[height=5.3cm,width=\linewidth,keepaspectratio]{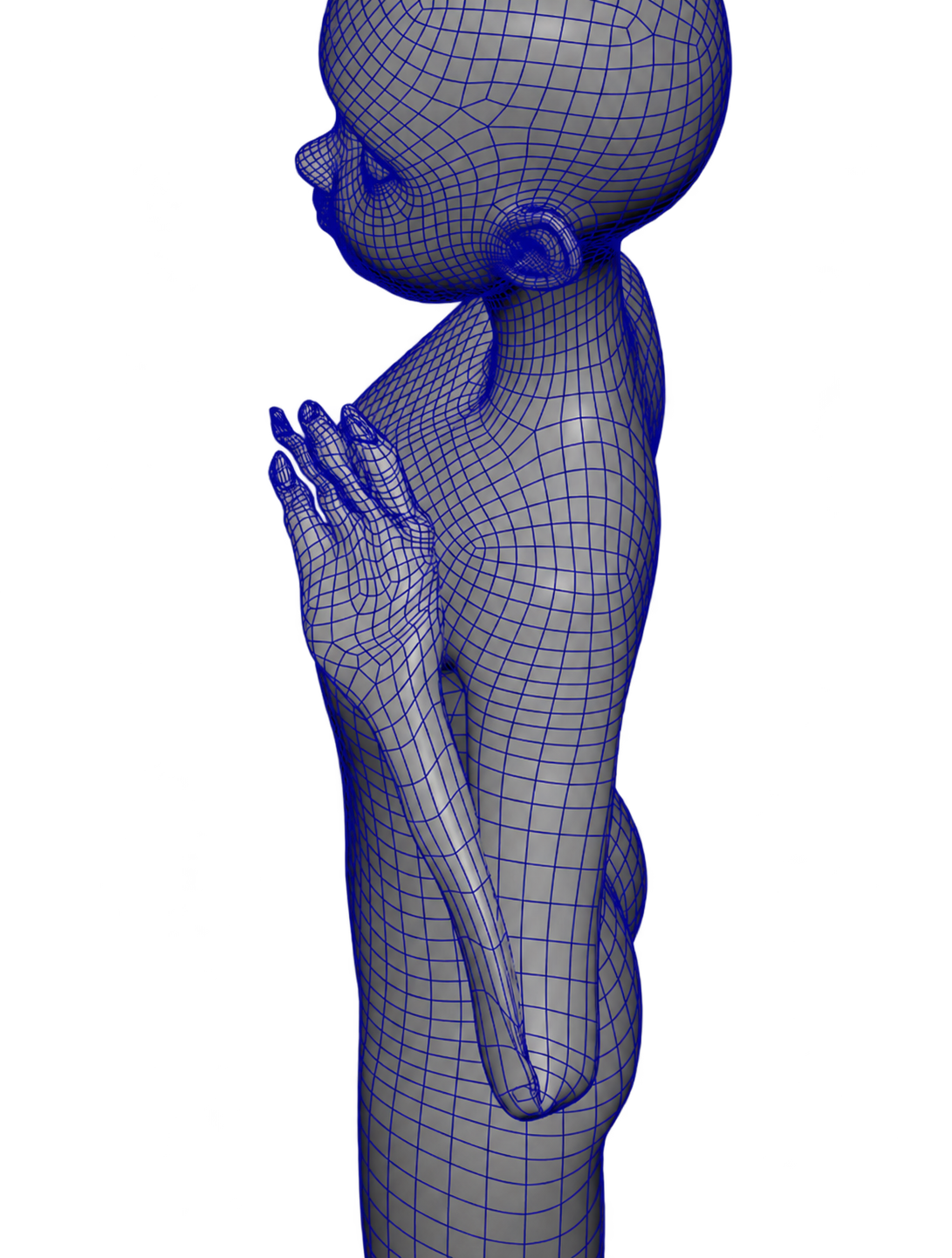}\\[3pt]
        \small (b) \textit{elbowSuperBentLf}
    \end{minipage}
    \caption{Neutral-to-pose predictions on the held-out character for two elbow poses.
    Each panel shows a predicted mesh reconstructed from a neutral source;
    neutral inputs and ground-truth targets are not displayed. The requested
    bending action is recognizable, while local distortion and pinching remain
    around the elbow.}
    \Description{Two side views show a predicted elbow bend and a stronger
    elbow bend, with wireframe overlays revealing local joint distortion.}
    \label{fig:elbow_neutral_to_pose}
\end{figure*}

The neutral-to-pose examples in Figure~\ref{fig:elbow_neutral_to_pose}
show predictions for \textit{elbowBentLf} and \textit{elbowSuperBentLf}.
Both produce a recognizable elbow bend, but visible surface distortion remains,
particularly around the joint. These outputs illustrate the distinction between
recovering the broad pose semantics and reconstructing a clean deformation.
The conditioning-strategy comparison for \textit{cornerUpLf} is shown in
Figure~\ref{fig:corner_conditioning}.

\begin{figure}[t]
    \centering
    \includegraphics[width=0.78\linewidth]{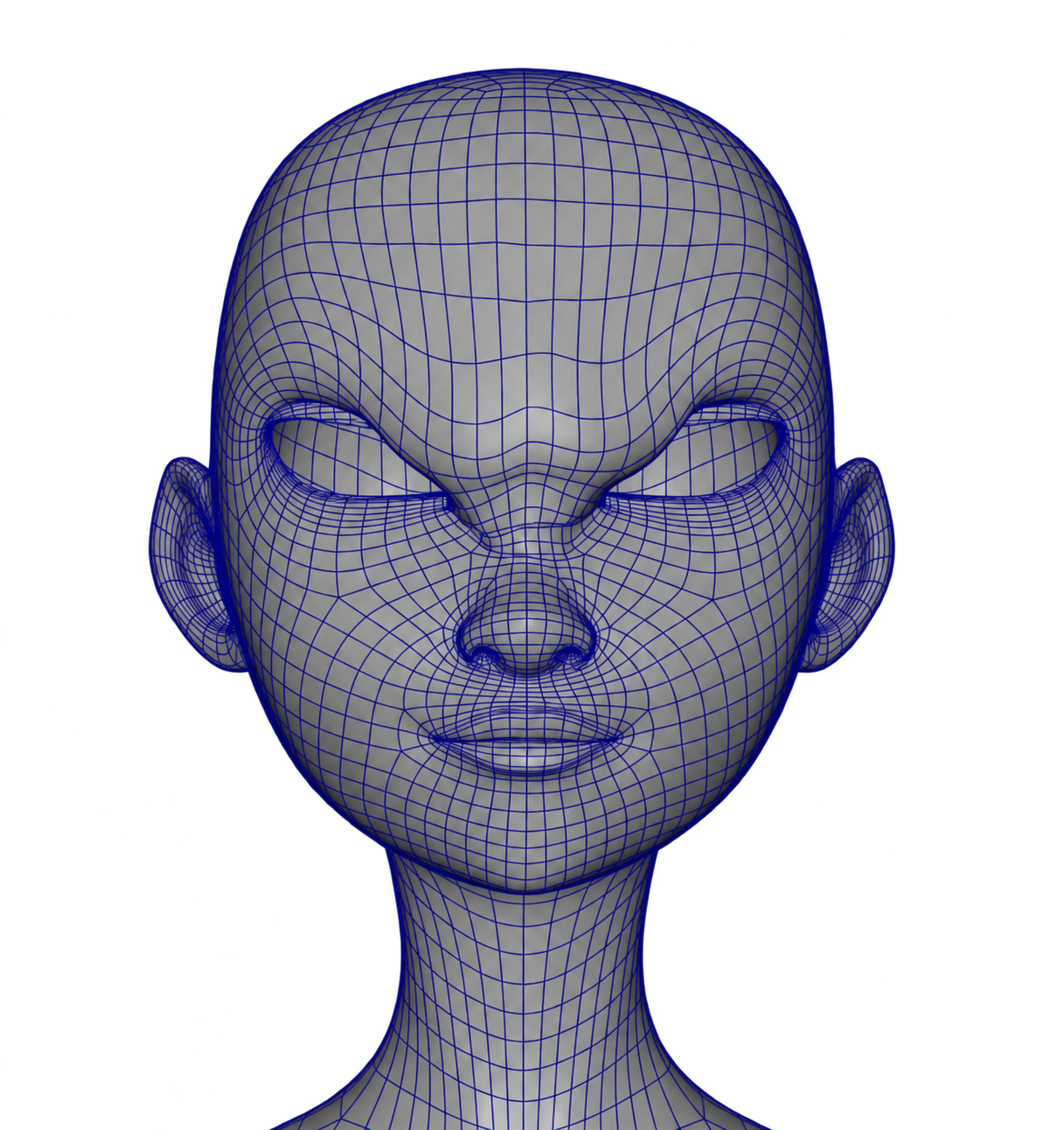}
    \caption{Combined \textit{browDwLf} and \textit{browDwRt} prediction on the held-out character.
    The bilateral brow-lowering action is visible, but the two sides are not
    reconstructed symmetrically. The panel shows the predicted combined state.}
    \Description{Frontal wireframe face with both brows lowered and visible left-right asymmetry.}
    \label{fig:brow_combined}
\end{figure}

A further example combines \textit{browDwLf} and \textit{browDwRt}
(Figure~\ref{fig:brow_combined}). Both brow regions respond to the combined
condition, but the predicted expression is asymmetric despite the intended
symmetric action. This illustrates a limitation in the consistency of the
reconstructed left and right components; the image alone does not isolate
whether the discrepancy originates in the latent condition or the decoder.

\begin{figure}[t]
    \centering
    \includegraphics[width=\linewidth]{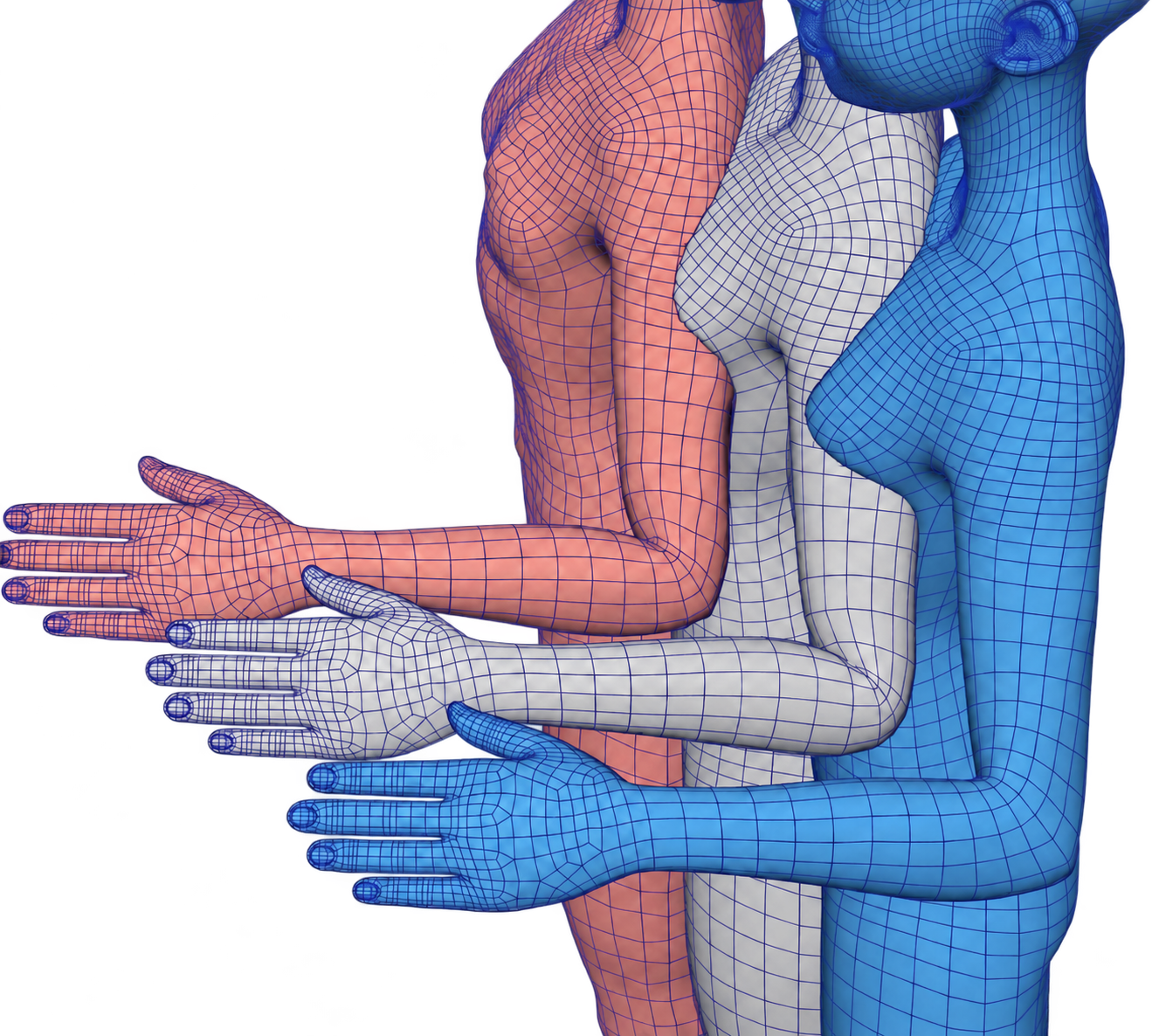}
    \caption{Corrective prediction for \textit{elbowBentLf} on the held-out character.
    Grey: skinning-only input. Red: LatentReRig prediction.
    Blue: ground-truth skinning-plus-blendshape target.
    The meshes are spatially offset for visual comparison.
    The predicted correction partially improves the elbow volume but remains
    weaker than the target and exhibits residual artifacts.}
    \Description{Three offset bent-arm meshes: red prediction, grey skinning-only source, and blue corrected ground truth.}
    \label{fig:elbow_corrective}
\end{figure}

Corrective-style prediction is evaluated using
\texttt{skinning\_\allowbreak to\_\allowbreak skinning\_\allowbreak plus\_\allowbreak blendshape}
pairs. Here, the broad skeletal motion is already present in the source mesh.
Figure~\ref{fig:elbow_corrective} compares the skinning-only input, the
predicted corrected mesh, and the artist-authored target for \textit{elbowBentLf}.
The prediction partially restores the elbow shape toward the target, but the
corrective effect is weaker and residual surface artifacts remain.

The prototype also reveals unintended motion outside the requested deformation
region. The knee and puff examples in Figure~\ref{fig:static_leakage} are
discussed in Section~\ref{sec:limitations}. Together, these examples expose
both recognizable pose responses and the remaining limitations of local
shape reconstruction and spatial isolation.

\section{Identity, Pose, and the Latent Noise Floor}
\label{sec:latent_analysis}

The mesh predictions in Section~\ref{sec:experiments} raise two questions
about the conditioning representation: whether it preserves character-specific
information, and whether pose-related changes are distinguishable from
SDF-resampling variability. The following diagnostic analyses examine these
properties to help interpret the observed transfer behaviour.


\subsection{Identity Separation}
\label{sec:identity_separation}

Identity information is evaluated using deterministic latent means computed
from neutral character states. For two identities $i$ and $j$, the
standardized latent distance is defined as

\begin{equation}
d_{\mathrm{id}}(i,j)
=
\left\|
\tilde{\mu}_i^0
-
\tilde{\mu}_j^0
\right\|_2,
\label{eq:identity_distance}
\end{equation}

where $\tilde{\mu}$ denotes a latent code transformed using the same global
standardization statistics.

\begin{figure}[t]
    \centering
    \includegraphics[width=\linewidth]{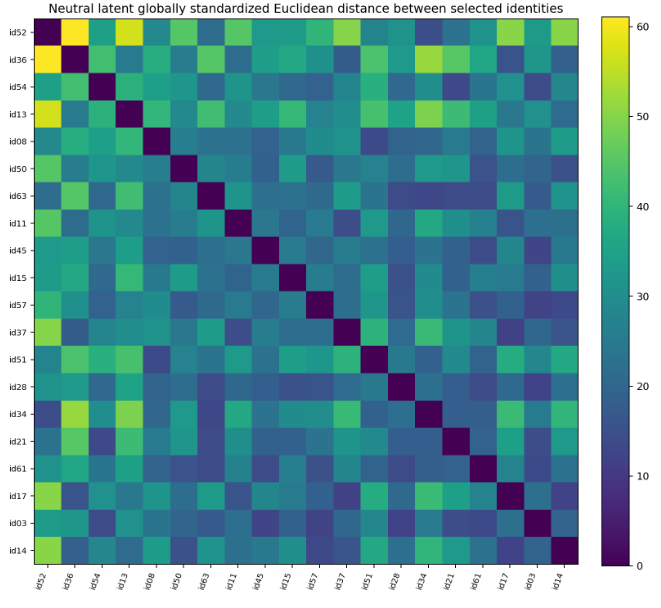}
    \caption{
        Pairwise standardized Euclidean distances between deterministic
        neutral latent means for 20 selected identities. The diagonal
        corresponds to zero self-distance, while the non-zero off-diagonal
        structure indicates separation between character identities.
    }
    \Description{Heatmap of pairwise standardized distances among neutral latent means for twenty character identities, with a zero diagonal.}
    \label{fig:neutral_identity_matrix}
\end{figure}

The resulting pairwise structure is shown in
Figure~\ref{fig:neutral_identity_matrix}. The non-zero off-diagonal
distances indicate that neutral character identities remain separated in the
standardized latent representation.

For the statistical analysis, 20 neutral identities are considered,
producing

\begin{equation}
\binom{20}{2} = 190
\end{equation}

unique inter-identity distances.

The mean inter-identity distance is approximately $27.01$; the full
descriptive statistics are reported in Table~\ref{tab:identity_noise_stats}.
These distances indicate that the representation retains differences between
characters. Their scale is assessed against repeated encodings of the same
geometry in the following analysis.

\begin{figure}[t]
    \centering
    \includegraphics[width=\linewidth]{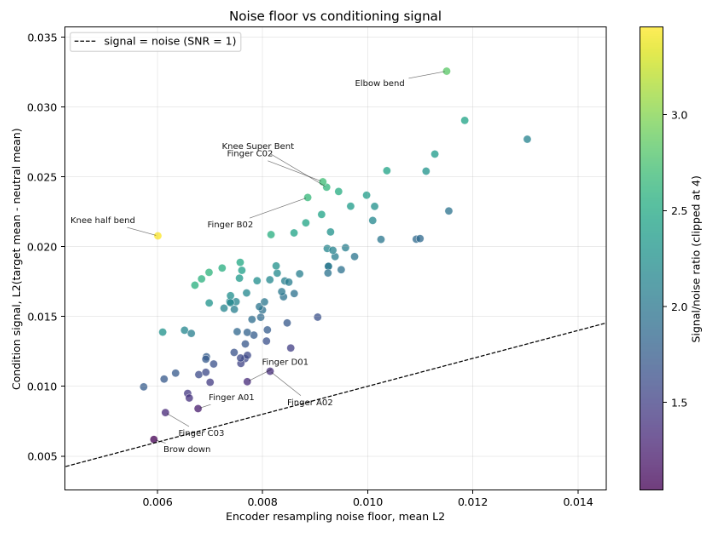}
    \caption{
        SDF-resampling variability versus pose-conditioning signal.
        The dashed line marks $R_{\mathrm{cond/noise}}=1$; points farther above
        the line exhibit a more clearly distinguishable conditioning signal.
    }
    \Description{Plot comparing pose-conditioning distances with the empirical same-geometry SDF-resampling baseline.}
    \label{fig:noise_floor}
\end{figure}

\begin{figure*}[!t]
    \centering
    \includegraphics[width=\textwidth]{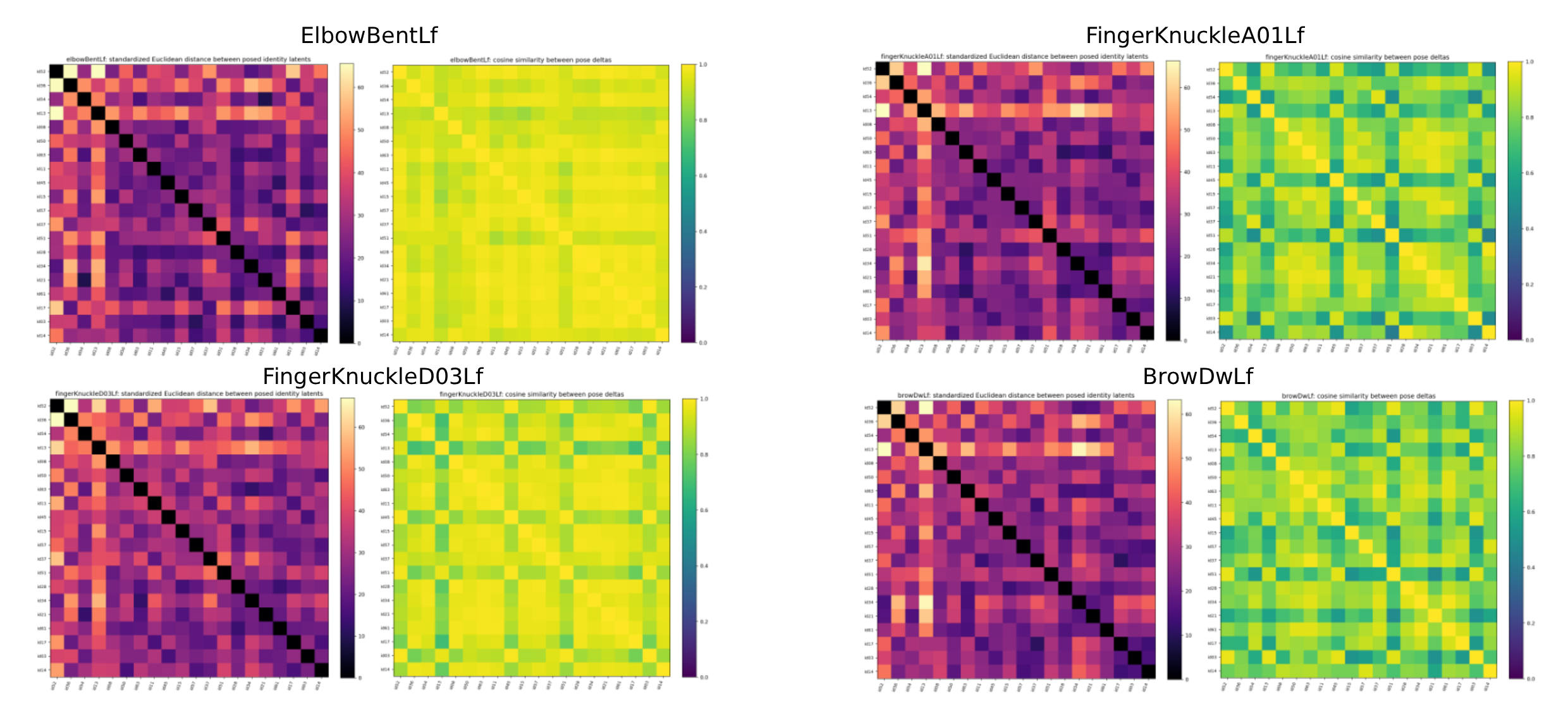}
    \caption{
        Identity--pose interaction for four representative semantic poses.
        For each pose, the left matrix reports pairwise standardized Euclidean
        distances between posed identity latents, while the right matrix reports
        cosine similarity between the corresponding neutral-to-pose latent deltas.
    }
    \Description{Four pairs of matrices compare identity distances and pose-delta cosine similarities for elbow, finger, and brow poses.}
    \label{fig:pose_entanglement}
\end{figure*}

\subsection{Encoder Noise Floor}
\label{sec:noise_floor}

The SDF encoder receives a finite spatial sampling rather than the complete
continuous field. Re-encoding the same geometry with a different sampling therefore
produces slightly different latent values. For two independent resamplings of mesh
state $m$, using the deterministic posterior mean for each encoding,

\begin{equation}
d_{\mathrm{noise}}
=
\left\|
\tilde{\mu}_m^{(a)}
-
\tilde{\mu}_m^{(b)}
\right\|_2.
\label{eq:noise_distance}
\end{equation}

This variability reflects sensitivity to the sampled SDF input, not sampling
from the variational posterior. The posterior mean is deterministic for a
fixed input and encoder in evaluation mode; different SDF samplings can still
produce different means. Repeated-encoding statistics are used only for
diagnosis, not to replace the latent inputs during training or inference.

A set of 190 same-geometry resampling distances has mean $8.30$. Compared
with the mean inter-identity distance of $27.01$, this gives

\begin{equation}
\frac{
\overline{d}_{\mathrm{identity}}
}{
\overline{d}_{\mathrm{noise}}
}
\approx
3.26.
\label{eq:identity_noise_ratio}
\end{equation}

Here $\overline{d}$ denotes a sample mean of distances, distinct from the
posterior mean $\mu$. The descriptive statistics are summarized in
Table~\ref{tab:identity_noise_stats}.

\begin{table}[t]
    \centering
    \caption{Inter-identity and noise-only latent distances.}
    \label{tab:identity_noise_stats}
    \begin{tabular}{lrr}
        \toprule
        Statistic & Identity & Noise \\
        \midrule
        $N$      & 190   & 190  \\
        Mean     & 27.01 & 8.30 \\
        Median   & 25.29 & 8.03 \\
        Std.     & 9.66  & 2.56 \\
        Min.     & 11.70 & 2.36 \\
        Max.     & 61.08 & 18.33 \\
        \bottomrule
    \end{tabular}
\end{table}

The same comparison is applied to deformation conditioning. A larger pose
signal relative to same-state resampling variability indicates a greater
separation from this empirical baseline. We define

\begin{equation}
R_{\mathrm{cond/noise}}
=
\frac{
d_{\mathrm{condition}}
}{
d_{\mathrm{noise}}+\varepsilon
}.
\label{eq:cond_noise_ratio}
\end{equation}

Across single-pose experiments, the mean ratio is approximately $2.03$ and the
median $2.06$. Strong knee, elbow, and clavicle conditions approach or exceed
$3$, while several brow, finger, knee, clavicle, and mouth-corner conditions
fall near $1.2$--$1.5$.

Figure~\ref{fig:noise_floor} therefore shows that most pose signals rise above
the empirical resampling baseline, but with substantially different margins.
These ratios help interpret representation-level sensitivity alongside
mesh-prediction errors. They are diagnostic summaries, rather than an absolute
threshold below which a deformation cannot be represented.

\subsection{Kolmogorov--Smirnov Analysis}
\label{sec:ks}

The Kolmogorov--Smirnov (KS) statistic summarizes the separation between
the empirical inter-identity and same-geometry resampling distance distributions.
The corresponding two-sample hypotheses are

\begin{equation}
\begin{aligned}
H_0 &: F_{\mathrm{identity}}(x)=F_{\mathrm{noise}}(x),\\
H_1 &: F_{\mathrm{identity}}(x)\neq F_{\mathrm{noise}}(x).
\end{aligned}
\label{eq:ks_hypotheses}
\end{equation}

For 190 distances in each group, the original two-sample calculation returned
\begin{equation}
D=0.9368,\qquad p_{\mathrm{nominal}}=3.15\times10^{-91}.
\end{equation}

The large value of $D$, together with the approximately $3.26$ ratio of mean
distances, describes strong separation in the observed data. However, the
190 inter-identity distances reuse 20 latent codes and are therefore not
independent observations. The nominal $p$-value does not account for this
dependence and is not interpreted as calibrated inferential significance.

This comparison measures separation relative to SDF-resampling variability.
It does not, on its own, establish semantic organization or disentanglement;
those questions also require the pose-consistency and transfer experiments.

\begin{figure*}[t]
    \centering
    \begin{minipage}[b]{0.58\textwidth}
        \centering
        \includegraphics[height=5.3cm,width=\linewidth,keepaspectratio]{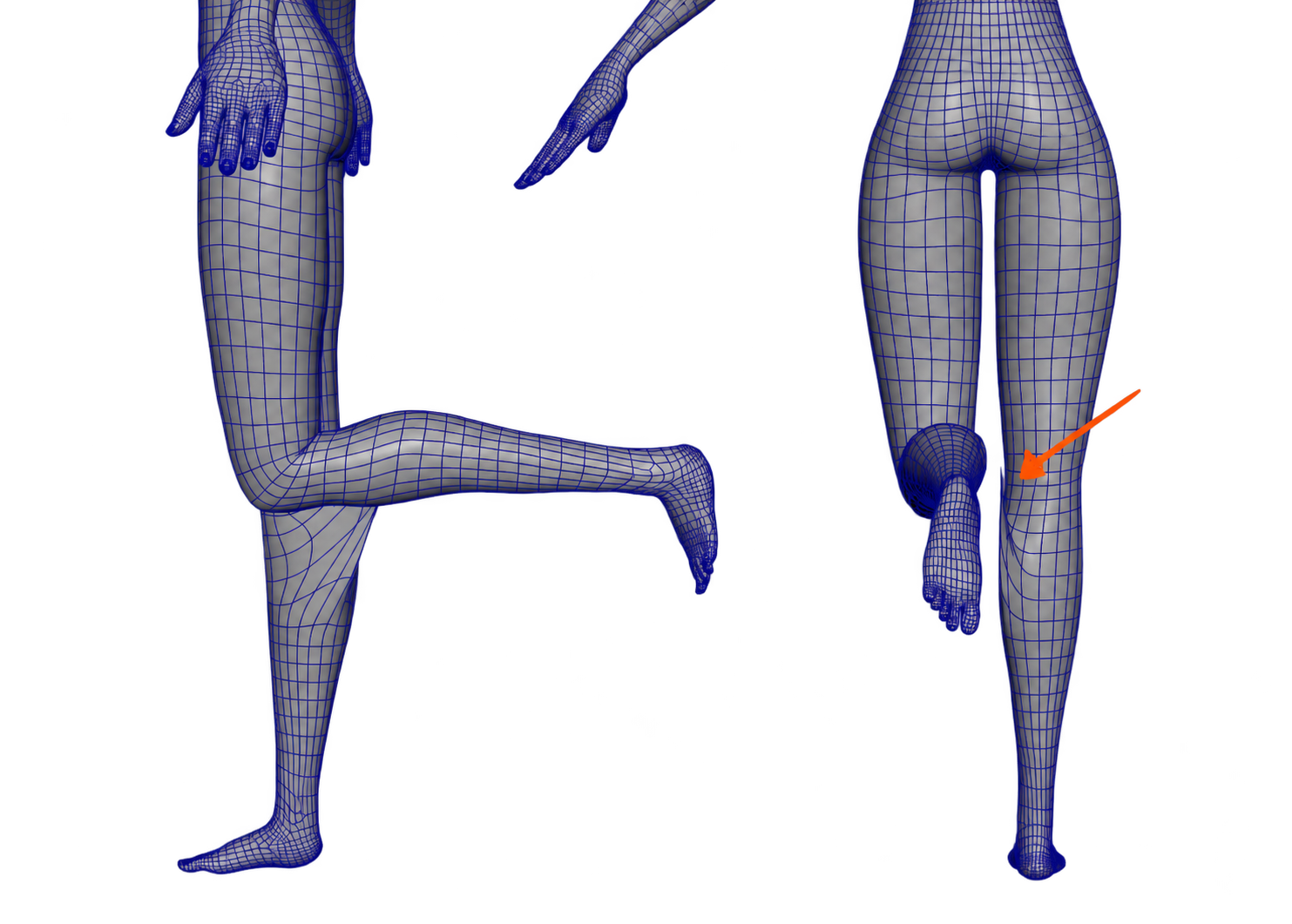}\\[3pt]
        \small (a) \textit{kneeBentLf}: side and front views
    \end{minipage}\hfill
    \begin{minipage}[b]{0.38\textwidth}
        \centering
        \includegraphics[height=5.3cm,width=\linewidth,keepaspectratio]{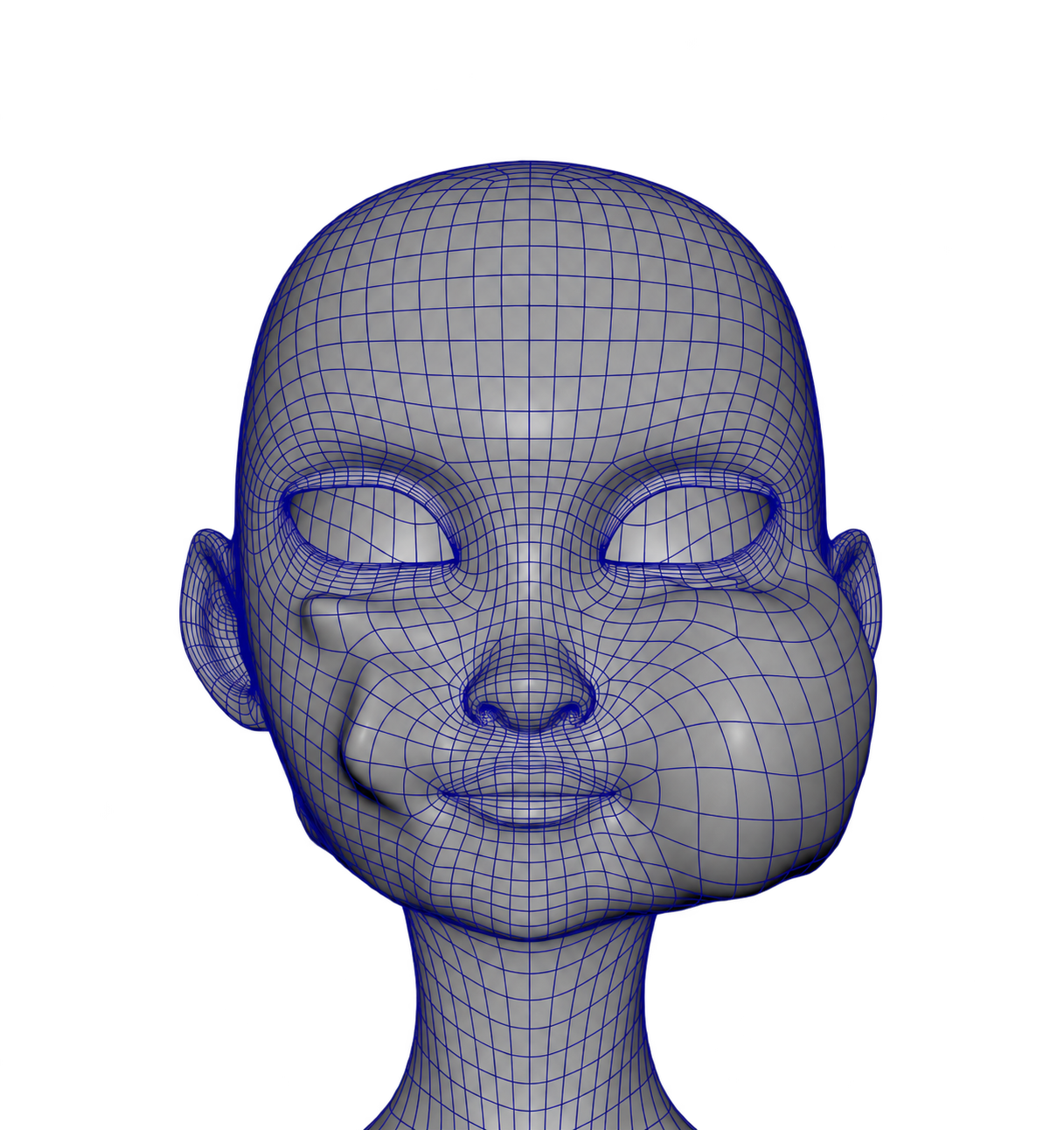}\\[3pt]
        \small (b) \textit{puffLf}
    \end{minipage}
    \caption{Static leakage in body and facial predictions on the held-out character.
    (a) The left-knee bend is recognizable, but the opposite leg is also
    deformed; the orange arrow marks an affected region that should remain
    static. (b) The left puff action is accompanied by unwanted movement on
    the opposite side of the face. Static leakage denotes predicted motion
    outside the region intended to move.}
    \Description{Side and front views of a knee-bend prediction, with an
    orange arrow indicating the opposite leg, alongside a frontal puff
    prediction with deformation extending to the opposite cheek.}
    \label{fig:static_leakage}
\end{figure*}

\subsection{Identity--Pose Entanglement}
\label{sec:identity_pose_entanglement}

Identity separation and pose consistency describe complementary properties:
absolute codes distinguish characters, while relative directions describe
how similarly they encode an analogous pose.

For pose $p$, identity structure is measured by

\begin{equation}
d_{\mathrm{id}}^p(i,j)
=
\left\|
\tilde{\mu}_i^p
-
\tilde{\mu}_j^p
\right\|_2,
\label{eq:posed_identity_distance}
\end{equation}

while pose consistency is measured through the cosine similarity in
Equation~\ref{eq:pose_cosine}. Figure~\ref{fig:pose_entanglement} summarizes
four representative cases, with numerical values reported in
Table~\ref{tab:entanglement_summary}.

\begin{table}[t]
    \centering
    \caption{Representative identity--pose interaction.}
    \label{tab:entanglement_summary}
    \begin{tabular}{lrr}
        \toprule
        Pose & ID dist. & Delta cosine \\
        \midrule
        elbowBentLf            & 29.22 & 0.952 \\
        fingerKnuckleD03Lf     & 31.67 & 0.921 \\
        fingerKnuckleA01Lf     & 30.10 & 0.791 \\
        browDwLf               & 29.01 & 0.800 \\
        \bottomrule
    \end{tabular}
\end{table}

Identity separation remains high in all four cases, while pose-direction consistency
varies substantially. \textit{elbowBentLf} is the clearest example: identities
remain separated, with a mean standardized distance of approximately $29.22$, while
the pose-delta cosine similarity reaches approximately $0.952$.
\textit{fingerKnuckleD03Lf} shows a similar pattern at $31.67$ and $0.921$.

The weaker \textit{fingerKnuckleA01Lf} and \textit{browDwLf} conditions preserve
identity structure but reduce mean delta similarity to approximately $0.79$--$0.80$.
Loss of pose consistency therefore does not correspond to identity collapse; rather,
the pose direction becomes more dependent on character-specific geometry.

These observations are consistent with residual identity--pose entanglement:
identity remains represented in the absolute codes, while the reusability of
relative directions depends on the pose.

\section{Discussion and Limitations}
\label{sec:limitations}

The experiments support the existence of reusable deformation structure, but also
show that latent organization and successful geometric prediction are distinct
problems. The results suggest two principal failure regimes.

In a \textit{conditioning failure}, the latent representation does not encode the
requested deformation strongly or consistently enough. This is most relevant for
small facial, finger, and corrective-style changes whose condition-to-noise ratio
approaches the empirical resampling baseline. Improving the explicit decoder alone
would be unlikely to solve these cases. The behaviour is compatible with known
information-capacity effects in variational models
\citep{burda2016iwae,alemi2018broken,higgins2017betavae} and, more generally, with
the risk that weak information is under-represented
\citep{bowman2016sentences,razavi2019deltavae}. The present results are better
interpreted as local or regime-specific under-representation than as complete
posterior collapse.

In a \textit{decoding failure}, by contrast, a coherent latent condition is available
but is not translated into sufficiently accurate geometry. Several broad poses fall
into this category: the requested anatomical region moves in the expected direction,
yet the output may still show under-expression, static leakage, local surface noise,
over-smoothing, or an incorrect distribution of deformation. Semantic plausibility
therefore does not imply production-level surface accuracy.

Static leakage is illustrated in Figure~\ref{fig:static_leakage}.
For \textit{kneeBentLf}, the requested leg bends, while the opposite leg,
which should remain static, is also affected. For \textit{puffLf}, the intended
cheek inflation is accompanied by unwanted deformation on the opposite side
of the face. These examples show that a recognizable semantic response can
coexist with inadequate spatial isolation of the predicted motion.

A further limitation is residual identity dependence. The occasional advantage of
\texttt{nearest\_\allowbreak identity\_\allowbreak delta} over the global mean shows that a semantic pose
direction is not fully identity-invariant. This should not necessarily be considered
undesirable. Character morphology and proportions affect how the same pose should
be realized geometrically; the useful objective is therefore a shared semantic
direction with identity-aware adaptation, rather than complete factorization.

The current representation also has two structural limits. Although SDF conditioning
reduces direct dependence on polygon connectivity, the explicit decoder still predicts
vertices on the source topology, so LatentReRig is not fully topology-independent.
Moreover, evaluation is restricted to a controlled humanoid-biped domain and does
not establish generalization to arbitrary character classes or unconstrained
production data.

\section{Conclusion and Future Work}
\label{sec:conclusions}

LatentReRig couples SDF-based variational representation learning with
explicit vertex prediction to investigate deformation transfer across
character identities. Several semantic poses produce coherent cross-identity
latent directions, and these signals can guide held-out characters toward
the intended deformation. Accurate mesh reconstruction remains less reliable,
particularly for small localized changes and corrective contributions.

The experiments identify complementary priorities for further development:
more stable conditioning for weak deformations and more accurate geometric
interpretation of the available signal. Future work should combine the global
SDF representation with local geometric features and extend evaluation across
more diverse identities, subtle correctives, and combined poses. The goal is
to preserve the reusability of latent deformation signals while improving the
precision and local control of the resulting mesh predictions.

\bibliographystyle{ACM-Reference-Format}
\bibliography{references}

\end{document}